\pdfoutput=1
\documentclass[10pt,twocolumn,letterpaper]{article}

\usepackage[pagenumbers]{cvpr} 

\usepackage[accsupp]{axessibility}
\usepackage{graphicx}
\usepackage{amsmath}
\usepackage{color}
\usepackage[table,xcdraw]{xcolor}
\usepackage{multirow}
\usepackage{amssymb}
\usepackage{threeparttable}
\usepackage{booktabs}
\usepackage{algorithm}
\usepackage{algorithmicx}
\usepackage{algpseudocode}

\usepackage[pagebackref,breaklinks,colorlinks]{hyperref}

\usepackage[capitalize]{cleveref}
\crefname{section}{Sec.}{Secs.}
\Crefname{section}{Section}{Sections}
\Crefname{table}{Table}{Tables}
\crefname{table}{Tab.}{Tabs.}

\def\cvprPaperID{7} 
\def\confName{CVPR}
\def\confYear{2023}

\begin{document}

\title{K-means Clustering Based Feature Consistency Alignment  \\ for Label-free Model Evaluation}

\author{Shuyu Miao, Lin Zheng, Jingjing Liu, and Hong Jin\\
Tiansuan Lab, Ant Group\\
{\tt\small \{miaoshuyu.msy, zhenglin.zhenglin, jingjing.lqq,  jinhong.jh\}@antgroup.com}
}
\maketitle

\begin{abstract}
The label-free model evaluation aims to predict the model performance on various test sets without relying on ground truths. The main challenge of this task is the absence of labels in the test data, unlike in classical supervised model evaluation. 
This paper presents our solutions for the 1st DataCV Challenge of the Visual Dataset Understanding workshop at CVPR 2023. Firstly, we propose a novel method called K-means Clustering Based Feature Consistency Alignment (KCFCA), which is tailored to handle the distribution shifts of various datasets. KCFCA utilizes the K-means algorithm to cluster labeled training sets and unlabeled test sets, and then aligns the cluster centers with feature consistency. Secondly, we develop a dynamic regression model to capture the relationship between the shifts in distribution and model accuracy. Thirdly, we design an algorithm to discover the outlier model factors, eliminate the outlier models, and combine the strengths of multiple autoeval models. On the DataCV Challenge leaderboard, our approach secured 2nd place with an RMSE of 6.8526. Our method significantly improved over the best baseline method by 36\% (6.8526 vs. 10.7378). Furthermore, our method achieves a relatively more robust and optimal single model performance on the validation dataset.

\end{abstract}

\section{Introduction}
\label{sec:intro}

\begin{figure}
	\centering
	\includegraphics[width=1\linewidth]{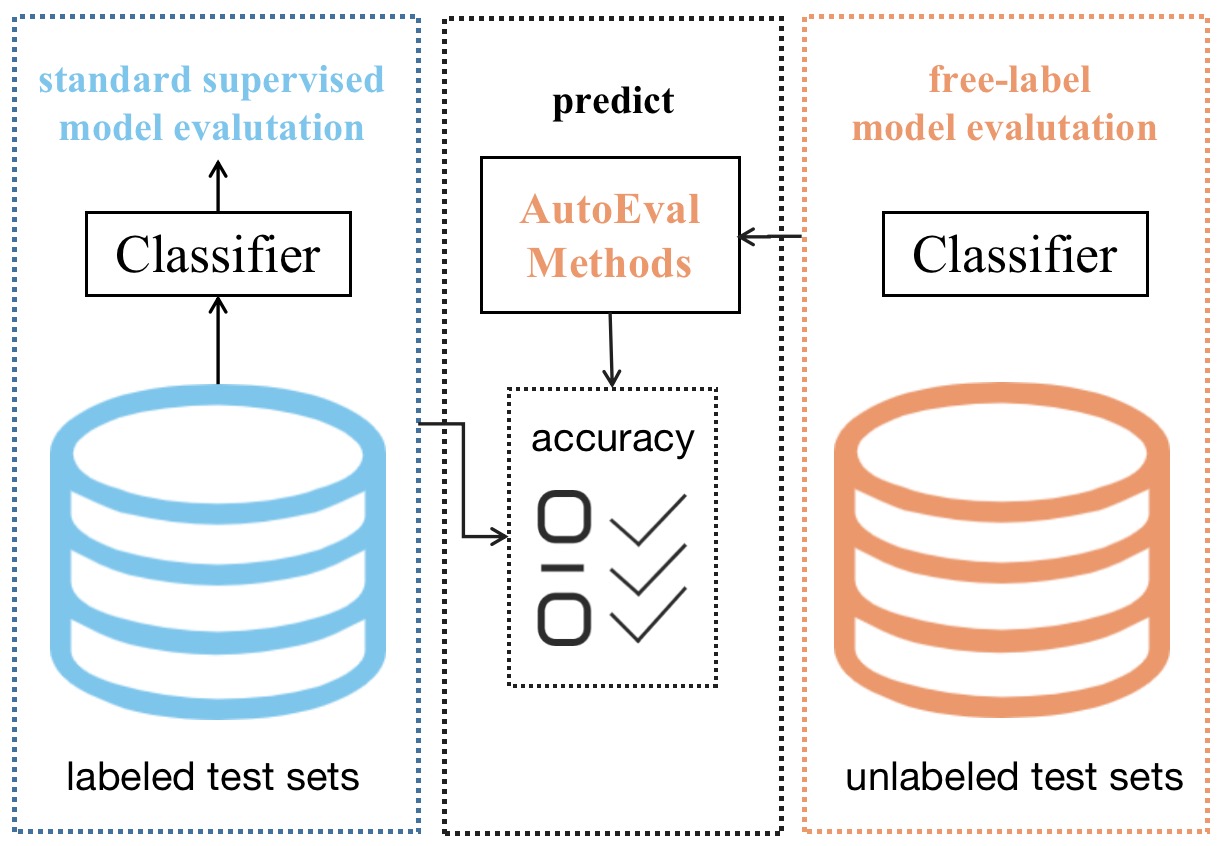}
	\caption{ The illustration of label-free model evaluation. Given a classifier trained on the training set, we can obtain its performance by evaluating it on labeled test data set, as shown in (left). However, in label-free model evaluation scenarios, we encounter unlabeled test data sets, and cannot use common metrics to evaluate our classifiers (right).
	}
	\label{fig-1}
\end{figure}

 Label-free model evaluation  task, also known as AutoEval \cite{deng2021labels}, requires models to evaluate the performance of datasets autonomously without explicit labels or categories. The models must identify inherent patterns and structures within the data without relying on pre-defined labels. Unlike supervised model evaluation \cite{deng2009imagenet, lin2014microsoft,miao2022balanced,pascal-voc-2012, DBLP:journals/corr/HuangLW16a,miao2021disentangled}, AutoEval does not require a vast amount of labeled data, as shown in Figure \ref{fig-1}, saving time and expensive costs. Furthermore, it can reveal potential data patterns and relationships that may not be discovered by supervised evaluation. However, this task is challenging due to the lack of explicit labels. Additionally, a test set comprises numerous images, and each image has varied and rich visual content \cite{deng2009imagenet}. In the 1st DataCV Challenge of the Visual Dataset Understanding workshop held at CVPR 2023 \cite{DataCVChallenge}, participants are required to design a model that can estimate the accuracy of a given model on test sets without ground truths.

In our daily lives, AutoEval mirrors real-world scenarios more closely. Evaluating the performance of an online model on out-of-time or out-of-distribution datasets typically requires data annotation, which can be prohibitively expensive and time-consuming. For instance, various risk data are often encountered in financial risk control scenarios. In order to detect various risky transactions, it is essential to evaluate the model’s performance in real-time. Hence, determining how to evaluate the model’s performance with unlabeled test datasets is crucial.

Recently, several studies have demonstrated promising performance in this task \cite{guillory2021predicting,deng2021labels,deng2021does,chen2021mandoline,jiang2021assessing,hendrycks2016baseline,gargleveraging}. 
Calibration generated on the unseen distribution (target domain) yields consistent estimates and further helps infer the model’s performance \cite{jiang2021assessing, gargleveraging}. However, methods that require calibration in the target domain frequently produce poor estimates because deep learning models trained and calibrated on seen data (source domain) may not be calibrated in the previously unseen target domain.
Some proposed methods \cite{deng2021labels, deng2021does, guillory2021predicting} introduce additional labeled data from several target domains to learn a regression function of a distributional distance, which then predicts model performance. This method assumes that a strong correlation exists between the invisible test set and the visible train/val set in the fundamental distance measurement.
The challenge baselines follow this paradigm, making it crucial to identify an appropriate linear correlation between the seen train sets and the unseen test sets and design an appropriate regression model. We address this challenge from three aspects: (1) designing excellent autoeval methods; (2) selecting the appropriate regressor; and (3) constructing the best integration strategy for multiple autoeval models.

To address the challenges stated above, we suggest three corresponding solutions.  Firstly, we propose a novel model, K-means Clustering Based Feature Consistency Alignment (KCFCA), capable of representing the distribution shifts in various datasets.
KCFCA utilizes the k-means clustering algorithm \cite{hartigan1979algorithm} to cluster the seen training set and unseen test set into clusters with a known number of categories. 
If the task at hand is N-classified, the centers of training samples and test samples that are clustered into N clusters should show close-to-distribution consistency. 
The distribution shifts between the two clustered centers can be used to fit a model regression.
Secondly, experimental evidence has proved that different regression models will have a significant impact on the final result \cite{deng2021labels}.
Therefore, we create a dynamic regression model that takes advantage of different regression models to fit the relationship between the shifts and the model accuracy.
Thirdly, we design an outlier model factor discovery algorithm to eliminate outlier models and integrate the advantages of multiple autoeval models. In the course of this, we discover an interesting phenomenon: autoeval models based on various pre-trained models exhibit remarkable performance gaps.
Lastly, our experiments validate the effectiveness of our solutions, and our model achieves second place on the DataCV Challenge leaderboard with an RMSE of 6.8526. 

To summarize, this paper’s main contributions are as follows:
\begin{itemize}
\item We propose a novel method, K-means Clustering Based Feature Consistency Alignment (KCFCA), which can represent the distribution shifts in various datasets.
\item We construct a dynamic regression model that fits the relationship between the distribution shifts and model accuracy.
\item We design an outlier model factor discovery algorithm to eliminate outlier models and integrate the advantages of multiple autoeval models.
\end{itemize}

\section{ Related Work}
Our work intersects with multiple related lines of research that have seen significant progress in recent years. Therefore, in this section, we provide a summary of the most closely related works.

\subsection{Label-free Model Evaluation}
Label-free model evaluation aims to predict the accuracy of an unseen test set when the ground truth is not accessible \cite{guillory2021predicting,deng2021labels,deng2021does,chen2021mandoline,jiang2021assessing,hendrycks2016baseline,gargleveraging,yu2022predicting}. This area has recently garnered widespread attention in the research community. Deng et al. \cite{deng2021labels} constructed a meta-dataset by transforming original images into various forms, adopted feature statistics to capture the distribution of a sample dataset, and trained a regression model to predict model performance. The difference of confidence was proposed \cite{guillory2021predicting} to yield successful estimates of a classifier’s performance across different shifts and model architectures. Such models rely on additional labeled data from several target domains to learn a linear regression function. The Average Thresholded Confidence (ATC) \cite{gargleveraging} method trained a threshold on the model’s confidence to predict accuracy as the fraction of unlabeled examples for which model confidence exceeds the threshold.

 \begin{figure*}
	\centering
	\includegraphics[width=1\linewidth]{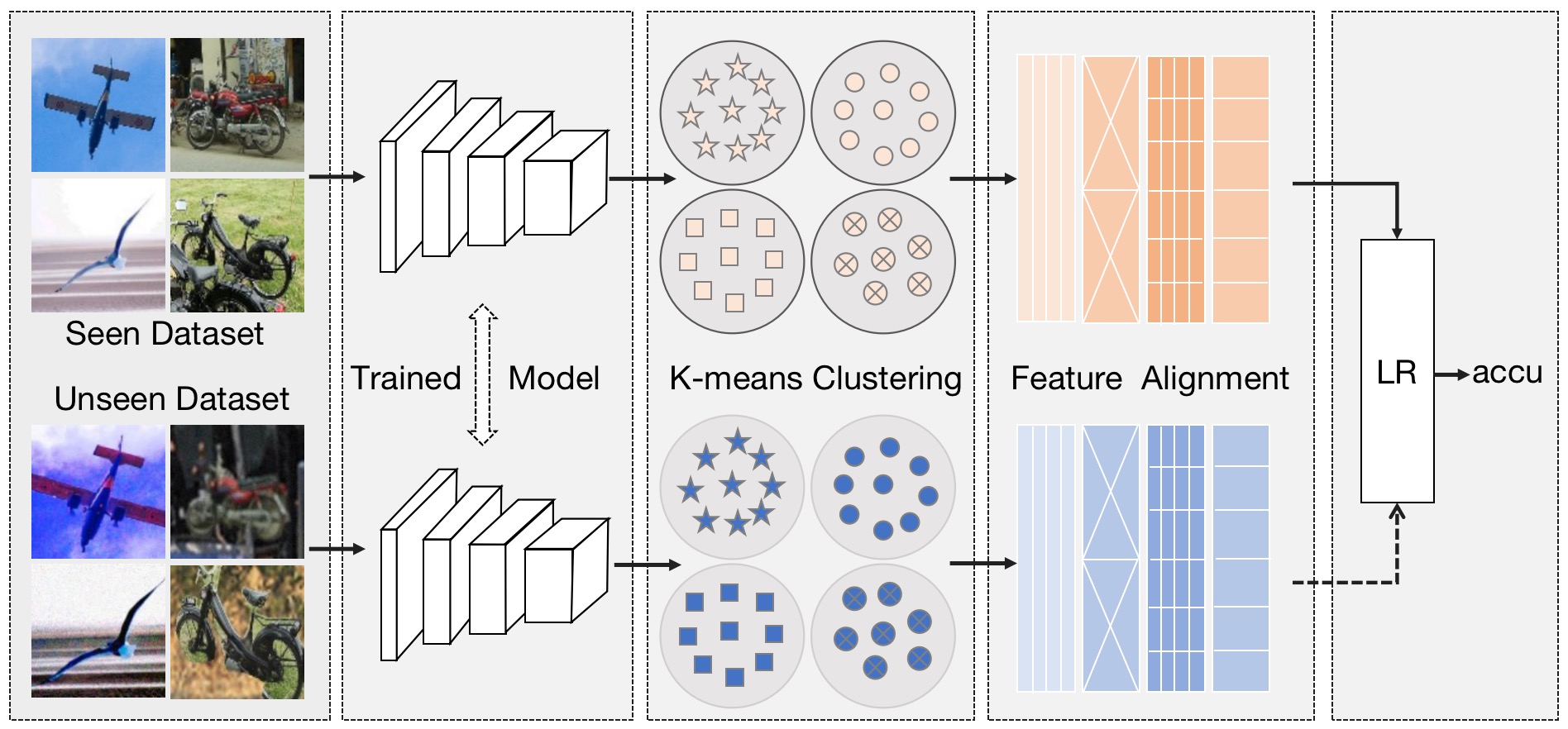}
	\caption{ The illustration of Clustering Based Feature Consistency Alignment (KCFCA). KCFCA performs feature extraction by a pre-trained model, feature clustering with the K-means algorithm, feature alignment of the clustered features, and finally regression of feature shift using a regression model.
	}
	\label{fig-overall-model}
\end{figure*}

\subsection{Out-Of-Distribution Detection}
Out-of-distribution (OOD) detection is a critical task in machine learning that aims to identify examples outside the realm of the training distribution \cite{ovadia2019can,ji2021predicting,hendrycksbaseline,geifman2017selective,liang2017enhancing,devries2018learning,ren2019likelihood,yu2022predicting,liu2020energy}. Model confidence outputs are commonly used as indicators to identify out-of-distribution samples \cite{hendrycksbaseline, geifman2017selective}. Liang et al. \cite{liang2017enhancing} proposed using temperature scaling and input perturbations to enhance OOD detection with model confidence. Devries and Taylor \cite{devries2018learning} introduced a method to learn confidence estimates for neural networks to produce intuitive and interpretable outputs. Sun et al. \cite{NEURIPS2021_01894d6f} designed ReAct - a straightforward and effective technique to reduce model overconfidence on OOD data, motivated by novel analysis on the internal activations of neural networks. Ren et al. \cite{ren2019likelihood} explored deep generative model-based approaches for OOD detection and observed that the likelihood score is heavily influenced by population-level background statistics. Learning the prediction uncertainty on OOD data remains a fundamental challenge in this task \cite{ovadia2019can, ji2021predicting}.

\subsection{Model Generalization Prediction}
Predicting the generalization capabilities of models \cite{arora2018stronger, corneanu2020computing, jiang2018predicting, neyshabur2017exploring, pmlr-v119-yang20j, schiff2021predicting, chen2020more, chen2018closing} on unseen data has been a topic of interest in research for a long time. Complexity measurements on trained models and training sets have been explored as a means of predicting the generalization gap \cite{arora2018stronger, corneanu2020computing, jiang2018predicting, neyshabur2017exploring}. Yang et al. \cite{pmlr-v119-yang20j} provided a simple explanation of this by measuring the bias and variance of neural networks to redefine how models generalize. Schiff et al. \cite{schiff2021predicting} used perturbation response (PR) curves to evaluate the accuracy change of a given network as a function of varying levels of training sample perturbation.
Our work focuses on the cluster difference concerning the prediction of unseen test sets.

\subsection{K-means Clustering}
The k-means clustering algorithm is a popular unsupervised machine learning algorithm that serves to partition a given dataset into k clusters \cite{hartigan1979algorithm, likas2003global, pham2005selection, kodinariya2013review, na2010research, hartigan1979k}. In this algorithm, each data point is assigned to the cluster whose centroid is closest to it. The algorithm iteratively updates cluster centroids until convergence, which is achieved when the assignment of data points to clusters no longer changes. The basic principle of the k-means algorithm is to minimize the sum of squared distances between each data point and its assigned cluster centroid. The algorithm randomly initializes k centroids and then assigns each data point to the nearest centroid. In the next step, the centroids are updated by computing the mean of all the data points assigned to each cluster. This process is repeated until convergence.
K-means is recognized as being both simple and efficient for partitioning datasets into k clusters. It has several advantages, such as computational efficiency and scalability.

\section{Methods}
\subsection{Problem Formulation}
We define this task by the source meta dataset $\mathcal{D}^m(\mathcal D_{train}, \mathcal D_{val})$ (i.e. seen training dataset), which consists of the labeled training data $\mathcal{D}_{train}$ and validation data $\mathcal{D}_{val}$. Following the challenge and approach in \cite{deng2021labels}, the source sample datasets $\mathcal{D}^s = \{\mathcal D_i(S_{xi}, S_{yi})\}^n _{i=1} $ are transformed from the original $\mathcal{D}^m$, where $S_{xi}$ is the \emph i-th training sample dataset, $S_{yi}$ are its corresponding labels, $n$ is the total number of the sample datasets, and $\mathcal D_i$ is the \emph i-th sample dataset. In addition, the target unlabeled test set is denoted as $\mathcal{D}_t$, and does not contain any ground truths. We assume that the model $\mathcal{M}(\theta_m)$ is pretrained on $\mathcal{D}^m$, where $\theta_m$ refers to the learned parameters that are fixed. If we have access to the label of $\mathcal{D}_t$, we can easily obtain the accuracy using $acc=\mathcal{M}(\mathcal{D}_t|\theta_m)$. However, in the absence of labeled ground truth, the objective of this task is to predict the accuracy of the unlabeled test set under the a priori conditions of $\mathcal{D}^m$ and $\mathcal{M}(\theta_m)$, 
which is expressed as Equation (\ref{equ_1}).
\begin{equation}
\label{equ_1}
    acc = f_\theta(\theta_\omega, \mathcal{D}_t|(\mathcal{D}^m,\mathcal{M}(\theta_m)))
\end{equation}
where $f_\theta(\cdot)$ represents the regression model that needs to be learned and $\theta_\omega$ are the parameters of this model.

Our work comprises three core components - K-means Clustering Based Feature Consistency Alignment (KCFCA), a novel AutoEval model that learns feature shifts between seen training data and unseen test data; Dynamic Regression Model (DRM), which strives to best fit the relationship between these shifts and model performance; and Outlier Model Factor Discovery (OMFD), which eliminates outlier autoeval models and integrates the advantages of multiple autoeval models. KCFCA will be discussed in Section \ref{KCFCA}, DRM in Section \ref{DRM}, and OMFD in Section \ref{OMFD}.

\subsection{K-means Clustering Based Feature Consistency Alignment }
\label{KCFCA}
If we assume the meta task is a $\mathcal N$ classification task, then $\mathcal{M}(\theta_m, X_i) \in 
\{ 1, 2, \cdots, \mathcal{N} \}$, where $X_i$ is \emph{i}-th input image. By using the k-means clustering algorithm to cluster sample features, we can theoretically divide them into $\mathcal{N}$ clusters with the best silhouette coefficient \cite{hartigan1979algorithm}.
Additionally, we propose K-means Clustering-based Feature Consistency Alignment, abbreviated as KCFCA and shown in Figure \ref{fig-overall-model}. We first review and summarize the K-means:
\begin{enumerate}
\item Initialize: Choose the number of clusters as K and select K random points (centroids) from the dataset as the initial centroids. 
\item Assign: Assign each data point to the nearest centroid based on the euclidean distance between the data point and the centroids.
\item Update: Recalculate the centroids of each cluster by taking the mean of all the data points in that cluster.
\item Iterate: Repeat steps 2 and 3 until convergence, which occurs when the centroids no longer change or a maximum number of iterations is reached.
\item Output: The algorithm outputs the K clusters and K cluster centers, where each cluster contains a set of data points that are similar to each other and dissimilar to data points in other clusters.
\end{enumerate}

As previously discussed, a dataset for a $\mathcal{N}$ classification task can be clustered into the most suitable $\mathcal{N}$ clusters. Given the sample datasets \{$\mathcal D_i$, $\mathcal D_{train}$, $\mathcal D_{val}$\}, and  pretrained model $\mathcal{M}(\theta_m)$, we construct dataset pairs as $\mathcal D_{sam} = \{\mathcal D_{sam_i} (\mathcal D_i, \mathcal D_{val})\}^n _{i=1}$ that can calculate the accuracy $acc_i$. For each pair $\mathcal D_{sam_i} = (\mathcal D_i, \mathcal D_{val})$, we first feed the $\mathcal D_i$ and $\mathcal D_{val}$ into the pretrained model $\mathcal{M}(\theta_m)$ to extract the feature map $\mathcal{F}_i$ and $\mathcal{F}_{val}$. The feature $\mathcal{F}_i$ and $\mathcal{F}_{val}$ are  then clustered into $\mathcal{N}$ clusters by K-means, with the cluster centers $\{\mathcal{F}_{ci}\}^{\mathcal{N}}_{ci=1}$ and $\{\mathcal{F}_{cval}\}^{\mathcal{N}}_{cval=1}$. Ideally, the distribution of each dataset pair should exhibit feature consistency. However, due to the uncertainty of unseen test data, distribution shifts often occur. Thus, we model the feature distance using frechet distance to fit these distribution shifts by \cite{dowson1982frechet,deng2021labels}. This distance is denoted as $D_i = D(\mathcal{F}_{ci}, \mathcal{F}_{cval})$. Finally, a regression model $f_\theta(\cdot)$ is designed to regress the relations between $D_i$ and $acc_i$.  KCFCA can be formulated as:
\begin{equation}
\label{equ_2}
    acc = f_\theta(\theta_\omega, D(\mathcal{K}(\mathcal{M}(\theta_m, \mathcal{D}_{val})), \mathcal{K}(\mathcal{M}(\theta_m, \mathcal{D}_{i})))
\end{equation}
where $\mathcal{K}(\cdot)$ is the k-means clustering algorithm.

The training and testing process of the model can be outlined as follows.
\begin{itemize}
    \item \textbf{\emph{Training}}: the regression model $f_\theta(\cdot)$ is adopted to learn the relation of $\{D_i, acc_i\}$.
    \item \textbf{\emph{Testing}}: calculate the feature distance between  $\mathcal{D}_{val}$ and $\mathcal{D}_{t}$, put it into the regression model $f_\theta(\cdot)$, and obtain the dataset accuracy. 
\end{itemize}
 Specifically, the KCFCA algorithm can be represented as the following  Algorithm \ref{alg-1}.

\begin{algorithm}
	\caption{K-means Clustering Based Feature Consistency Alignment }
	\label{alg-1}
	\begin{algorithmic}[1]
		\Require
		The sample datasets \{$\mathcal D_i$, $\mathcal D_{train}$, $\mathcal D_{val}$\}, and pretrained model $\mathcal{M}(\theta_m)$
		
		\Ensure		
		The accuracy of the unlabeled test set $\mathcal{D}_{t}$.
            \State 
              \textbf{Training:}
		\For{number of source sample datasets training:}	
		\State 
		a. Feature extraction 
   \State
    ~~~ $\mathcal{F}_i =  \mathcal{M}(\theta_m, \mathcal{D}_{i}), 
        \mathcal{F}_{val} =  \mathcal{M}(\theta_m, \mathcal{D}_{val})
        $
        \State b. K-means clustering
		\State 
              ~~~  $\{\mathcal{F}_{ci}\}^{\mathcal{N}}_{ci=1} = \mathcal{K}(\mathcal{F}_i), \{\mathcal{F}_{cval}\}^{\mathcal{N}}_{cval=1} = \mathcal{K}(\mathcal{F}_{val})$
		\State c. Feature distance
            \State 
            ~~~ $D_i = D(\mathcal{F}_{ci}, \mathcal{F}_{cval})$
		\State d. Learn regression model
            \State 
            $acc = f_\theta(\theta_\omega, D(\mathcal{K}(\mathcal{M}(\theta_m, \mathcal{D}_{val})), \mathcal{K}(\mathcal{M}(\theta_m, \mathcal{D}_{i})))$
		\EndFor
        \State 
        \State
        \textbf{Testing:}
        \State Calculate the accuracy of $\mathcal D_{test}$:
        \State 
        $acc = f_\theta(\theta_\omega, D(\mathcal{K}(\mathcal{M}(\theta_m, \mathcal{D}_{val})), \mathcal{K}(\mathcal{M}(\theta_m, \mathcal{D}_{t})))$
        \State \textbf{END}
    \end{algorithmic}
\end{algorithm} 

\subsection{Dynamic Regression Model}
\label{DRM}
\begin{figure}
	\centering
 \includegraphics[width=1.0\linewidth]{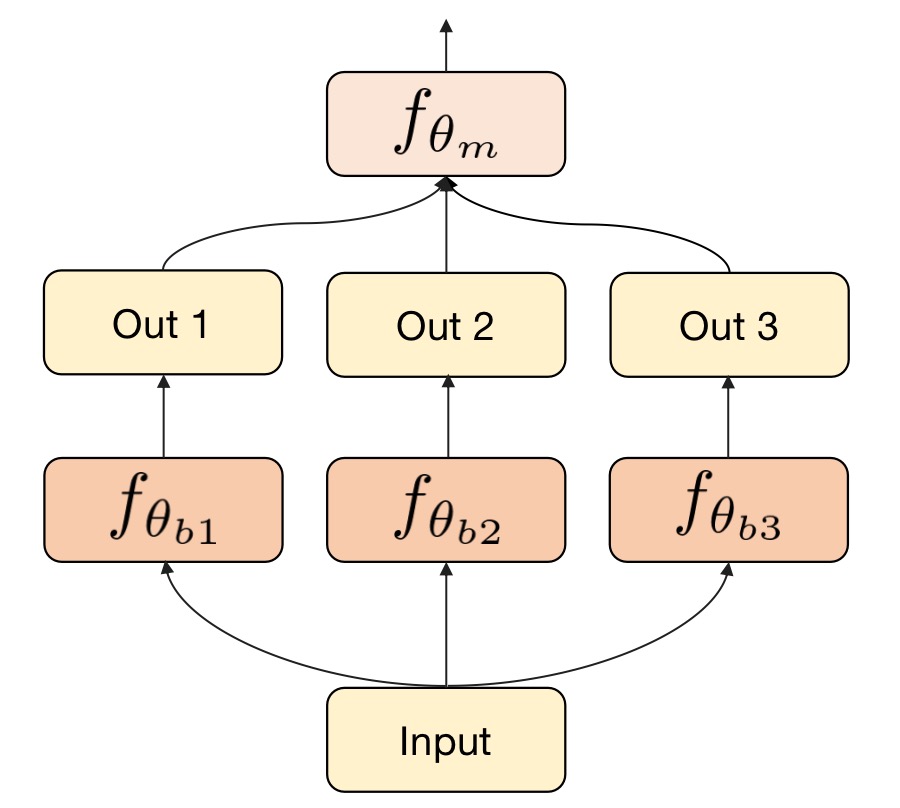}
	\caption{ The illustration of proposed DRM. $f_{\theta_{bm}}$ means the base regression models, and $f_{\theta_{m}}$ is the meta regression model.
	}
	\label{fig-model}
\end{figure}

In Section \ref{sec:intro} and in our experiments of Section \ref{Experiments_regre},  we observed that different regression models have distinct advantages when using the same feature input. This highlights the importance of designing a suitable regression model. To address this issue, we propose a Dynamic Regression Model, named DRM, that incorporates the advantages of multiple regression models. DRM comprises several base regressors $f_{\theta_{b}} = \{f_{\theta_{bi}}\}_{i=1}^m $ and a meta-regressor $f_{\theta_{m}}$, where $m$ is the number of base regression models. As shown in Figure \ref{fig-model}, the base regression models decouple to learn several sets of differentiated feature relations, while the meta-regression model dynamically fuses them to obtain a superior regression model. In other words, the meta-regressor learns base regression permutations of importance among the models. This procedure can be expressed as the following two equations: 
\begin{equation}
\label{equ_3}
    f_{\theta_{m}} = \{\omega_1, \omega_2, \cdots, \omega_m\}
\end{equation}
\begin{equation}
\label{equ_4}
    f_{\theta} = \omega_1 \cdot f_{\theta_{b1}} + \omega_2 \cdot f_{\theta_{b2}}  + \cdots +  \omega_m \cdot f_{\theta_{bm}}
\end{equation}
where $\omega_i$ corresponds to the weight of the \emph i-th base regressor.

For the base regression models, the following algorithms can be utilized in the specific implementation: 1) Linear Regression. It works by finding the line of best fit that describes the relationship between the input variables (also known as features) and the output variable (also known as the target variable); 2) K-Nearest Neighbors Regressor. It works by finding the k closest data points to a given input data point in the feature space and then taking the average (or median) of the output variable of those k data points;
3) Support Vector Regression. It uses support vector machines (SVMs) to find the hyperplane that best separates the data into different classes. 4) Random Forest Regressor. It is an ensemble-based algorithm that builds multiple decision trees on random subsets of the data and input variables, and then averages the predictions of each tree to make the final prediction. As for the meta-regression model, we can easily adopt a fully connected network or vote regression.

\subsection{Outlier Model Factor Discovery}
\label{OMFD}
As shown in Section \ref{Experiments_methods}, we observed that different autoeval algorithms have varying performance across distinct unlabeled test sets. In this challenge, combining different autoeval algorithms can improve the final prediction performance significantly. However, we found that simply fusing the results of all the algorithms cannot achieve optimal outcomes because outlier models may appear on different datasets or pre-trained models. Hence, we propose an Outlier Model Factor Discovery (OMFD) method to eliminate autoeval algorithms with lower performance stability.

Intuitively, most models predict that the consistency of results is more likely to be the correct result. Conversely, there is a possibility of the results being wrong. We denote the various performances of autoeval algorithms as $\mathcal{A} = \{\mathcal{A}_i\}_{i=1}^m$, where $\mathcal{A}_i$ is the \emph{i}-th autoeval model. We define a threshold $\tau$ for the anomaly factor that measures whether the model is an anomalous outlier. The flow of OMFD can be illustrated as:

\begin{table*}
\renewcommand\arraystretch{1.2}
\small
\begin{center}
\begin{tabular}{c|cccc|cccc}
\toprule[1.5pt]                        & \multicolumn{4}{c|}{ Dataset A (RMSE $\downarrow$)}                                                                                                                                                                                                                                    & \multicolumn{4}{c}{Dataset B (RMSE $\downarrow$)}                                                                                                                                                                                                             \\ \cline{2-9} 
\multirow{-2}{*}{\cellcolor[HTML]{FFFFFF}Method} & \multicolumn{1}{c}{{\color[HTML]{24292F} {CIFAR-10.1}}}           & \multicolumn{1}{c}{{\color[HTML]{24292F} {CIFAR-10.1-C}}} & \multicolumn{1}{c}{{\color[HTML]{24292F} {CIFAR-10-F}}} & {\color[HTML]{24292F} {Overall}}              & \multicolumn{1}{c}{{\color[HTML]{24292F} {CIFAR-10.1}}} & \multicolumn{1}{c}{{\color[HTML]{24292F} {CIFAR-10.1-C}}} & \multicolumn{1}{c}{{\color[HTML]{24292F} {CIFAR-10-F}}} & {\color[HTML]{24292F} {Overall}} \\ \midrule[1.5pt]
{\color[HTML]{24292F} ConfScore \cite{hendrycksbaseline}}                 & \multicolumn{1}{c}{\cellcolor[HTML]{FFFFFF}{\color[HTML]{262626} 2.190}} & \multicolumn{1}{c}{{\color[HTML]{24292F} 9.743}}                 & \multicolumn{1}{c}{{\color[HTML]{24292F} 2.676}}               & {\color[HTML]{24292F} 6.985}                         & \multicolumn{1}{c}{1.584}                                      & \multicolumn{1}{c}{9.897}                                        & \multicolumn{1}{c}{2.63}                                       & \cellcolor[HTML]{FFFC9E}7.074           \\ \hline
{\color[HTML]{24292F} Entropy \cite{guillory2021predicting}}                   & \multicolumn{1}{c}{\cellcolor[HTML]{FFFFFF}{\color[HTML]{262626} 2.424}} & \multicolumn{1}{c}{{\color[HTML]{24292F} 10.300}}                & \multicolumn{1}{c}{{\color[HTML]{24292F} 2.913}}               & {\color[HTML]{24292F} 7.402}                         & \multicolumn{1}{c}{1.849}                                      & \multicolumn{1}{c}{10.537}                                       & \multicolumn{1}{c}{2.949}                                      & 7.561                                   \\ \hline
{\color[HTML]{24292F} Rotation \cite{deng2021does}}                  & \multicolumn{1}{c}{{\color[HTML]{24292F} 7.285}}                         & \multicolumn{1}{c}{{\color[HTML]{24292F} 6.386}}                 & \multicolumn{1}{c}{{\color[HTML]{24292F} 7.763}}               & {\color[HTML]{24292F} 7.129}                         & \multicolumn{1}{c}{--}                                         & \multicolumn{1}{c}{--}                                           & \multicolumn{1}{c}{--}                                         & --                                      \\ \hline
{\color[HTML]{24292F} ATC \cite{gargleveraging}}                       & \multicolumn{1}{c}{{\color[HTML]{24292F} 11.428}}                        & \multicolumn{1}{c}{{\color[HTML]{24292F} 5.964}}                 & \multicolumn{1}{c}{{\color[HTML]{24292F} 8.960}}               & {\color[HTML]{24292F} 7.766}                         & \multicolumn{1}{c}{10.129}                                     & \multicolumn{1}{c}{7.131}                                        & \multicolumn{1}{c}{7.044}                                      & 7.178                                   \\ \hline
{\color[HTML]{24292F} FID \cite{deng2021labels}}                       & \multicolumn{1}{c}{{\color[HTML]{24292F} 7.517}}                         & \multicolumn{1}{c}{{\color[HTML]{24292F} 5.145}}                 & \multicolumn{1}{c}{{\color[HTML]{24292F} 4.662}}               & \cellcolor[HTML]{FFCC67}{\color[HTML]{24292F} 4.985} & \multicolumn{1}{c}{11.28}                                      & \multicolumn{1}{c}{5.683}                                        & \multicolumn{1}{c}{8.265}                                      & 7.258                                   \\ \hline
KCFCA (ours)                                            & \multicolumn{1}{c}{9.979}                                                & \multicolumn{1}{c}{8.828}                                        & \multicolumn{1}{c}{3.905}                                      & \cellcolor[HTML]{FFFC9E} 6.766                        & \multicolumn{1}{c}{13.223}                                     & \multicolumn{1}{c}{6.71}                                         & \multicolumn{1}{c}{6.562}                                      & \cellcolor[HTML]{FFCC67}6.876           \\ \bottomrule[1.5pt]
\end{tabular}
\caption{The validation dataset results are based on the same pre-trained model. The colored cells are sub-optimal and optimal performance. ``Dataset A'' indicates the training dataset is provided by the challenge; ``Dataset B'' denotes the training dataset is regenerated using the same transformed strategy as the challenge and \cite{deng2021labels}. Note that we just generate such a training dataset in this part to validate the robustness of various autoeval methods, and we just use the training dataset provided by the challenge to summit in the challenge leaderboard. }
	\label{table:methods}
 \end{center}
\end{table*}


\begin{enumerate}
\item Initialize: Visualize the performance of the autoeval model and manually select the appropriate centroid.
\item Calculate: Calculate the distance between the other autoeval models and this center.
\item Mark: If the maximum distance is greater than threshold $\tau$, the corresponding model is marked as an outlier.
\item Iterate: Repeat steps 1, 2 and 3 until convergence, which occurs when the maximum distance is no longer greater than the threshold $\tau$.
\item Output: Autoeval models marked as outliers.
\end{enumerate}
We blend all the autoeval models except outlier models to achieve the best model performance. Note that the threshold $\tau$ can be debugged based on the validation set or set empirically.


\section{Experiments}
\label{experiments}
In all of our experiments, we follow the same dataset and settings as the DataCV Challenge \cite{DataCVChallenge}.
\subsection{Experimental Settings}
\subsubsection{Datasets}
\begin{itemize}
    \item Training dataset: The training dataset consists of 1,000 transformed datasets from the original CIFAR-10 test set, using the transformation strategy proposed by Deng et al. \cite{deng2021labels}.
    \item Validation dataset:  The validation set was composed of CIFAR-10.1 \cite{recht2018cifar10.1,torralba2008tinyimages}, CIFAR-10.1-C \cite{hendrycks2019robustness} (add corruptions to CIFAR-10.1 dataset), and CIFAR-10-F (real-world images collected from Flicker\footnote{https://www.flickr.com/}.)
    \item Test dataset: The test set comprises 100 datasets\footnote{https://github.com/xingjianleng/autoeval\_baselines} provided by the challenge \cite{DataCVChallenge}.
\end{itemize}

\begin{table*}
\renewcommand\arraystretch{1.2}
\small
\begin{center}
\begin{tabular}{c|cccccc|cccccc}
			\toprule[1.5pt]
\cellcolor[HTML]{FFFFFF}                         & \multicolumn{6}{c|}{ResNet-56 (RMSE $\downarrow$) }                                                                                                                                                      & \multicolumn{6}{c}{RepVGG-A0 (RMSE $\downarrow$)}                                                                                                 \\ \cline{2-13} 
\multirow{-2}{*}{\cellcolor[HTML]{FFFFFF}Method} & {\color[HTML]{24292F} LR}                            & {\color[HTML]{24292F} KNN}   & {\color[HTML]{24292F} SVR}   & {\color[HTML]{24292F} MLP}  & {\color[HTML]{24292F} RFR}                           & DRM   & {\color[HTML]{24292F} LR} & {\color[HTML]{24292F} KNN} & {\color[HTML]{24292F} SVR} & {\color[HTML]{24292F} MLP} & {\color[HTML]{24292F} RFR}     & DRM    \\ 			\midrule[1.5pt]

{\color[HTML]{24292F} ConfScore \cite{hendrycks2016baseline}}                 & \cellcolor[HTML]{FFFFFF}{\color[HTML]{262626} 6.985} & {\color[HTML]{24292F} 7.708} & {\color[HTML]{24292F} 7.559} & {\color[HTML]{24292F} 12.028}     & {\color[HTML]{24292F} 7.765}                      & 7.503 & 8.721                     & 8.998                      & 9.647                      & \cellcolor[HTML]{FFFFFF}16.603 & 9.098 & 8.841  \\ \hline
{\color[HTML]{24292F} Entropy \cite{guillory2021predicting}}                   & \cellcolor[HTML]{FFFFFF}{\color[HTML]{262626} 7.401} & {\color[HTML]{24292F} 7.510} & {\color[HTML]{24292F} 8.033} & {\color[HTML]{24292F} 18.284} & 7.695 & 7.546 & 9.093                     & 9.398                      & 9.647                      & 9.647     & 9.566                     & 9.277  \\ \hline
{\color[HTML]{24292F} Rotation \cite{deng2021does}}                  & {\color[HTML]{24292F} 7.129}                         & {\color[HTML]{24292F} 7.723} & {\color[HTML]{24292F} 7.502} & {\color[HTML]{24292F} 13.207} & 8.209                         & 7.603 & 13.391                    & 11.144                     & 10.130                     & 18.651 & 11.303                        & 11.172 \\ \hline
{\color[HTML]{24292F} ATC \cite{gargleveraging}}                       & {\color[HTML]{24292F} 7.765}                         & {\color[HTML]{24292F} 6.700} & {\color[HTML]{24292F} 5.500} & {\color[HTML]{24292F} 13.202}   &  7.237                      & 6.578 & 8.132                     & 6.951                      & 5.806                      & 18.501  & 7.495                        & 6.561  \\ \hline
{\color[HTML]{24292F} FID \cite{deng2021labels}}                       & {\color[HTML]{24292F} 4.985}                         & {\color[HTML]{24292F} 5.825} & {\color[HTML]{24292F} 5.196} & {\color[HTML]{24292F} 19.508} & 5.330 & 5.273 & 5.965                     & 4.801                      & 4.583                      & 19.400     & 5.330                    & 4.703 \\ 			\toprule[1.5pt]

\end{tabular}
\caption{Experimental results  of different regression models on the same ``Overall" validate datasets. ``LR'' is LinearRegression, ``KNN'' is the KNeighborsRegressor, ``MLP'' is the MLPRegressor, ``RFR'' is the RandomForestRegressor, and ``DRM'' is our dynamic regression model.} 
\label{table:regression}
\end{center}
\end{table*}

\subsubsection{Pretrained classifier models}
In our experiments, we follow this challenge and evaluate the classifiers ResNet-56 \cite{he2016deep} and RepVGG-A0 \cite{ding2021repvgg}. Both implementations can be accessed in the public repository at the website\footnote{https://github.com/chenyaofo/pytorch-cifar-models}. To benefit from the models and load their pre-trained weights, use the code provided on the website.


\subsubsection{Evaluation metrics}
The evaluation metric used in our experiments is the root-mean-square error (RMSE), which can be formulated as:
\begin{equation}
\label{equ_6}
   RMSE = \sqrt{\frac{\sum_{t=1}^{T}\left ( \widehat{y}_t - y_t \right )^2}{T}}
\end{equation}
\subsection{Experiments and Findings}
To verify the effectiveness of our proposed KCFCA, DRM, and OMFD, we conducted detailed experiments on the same validation datasets.

\subsubsection{Experiments on various autoeval methods}
\label{Experimets_method_sin}
In our study, we conduct comprehensive experiments on various autoeval methods, including ConfScore \cite{hendrycksbaseline}, Entropy \cite{guillory2021predicting}, Rotation \cite{deng2021does}, ATC \cite{gargleveraging}, FID \cite{deng2021labels}, and our proposed KCFCA. All of the experiments are conducted based on the pre-trained ResNet56 provided by the challenge. Moreover, to validate the robustness of the various models, we perform experiments on 1,000 transformed datasets (denoted as  ``Dataset A") provided by the challenge and an additional 1,000 transformed datasets (denoted as ``Dataset B") generated using the same transformation strategy as the challenge and \cite{deng2021labels}. Note that we only use the training dataset provided by the challenge to submit the challenge results. 

\begin{table}
\renewcommand\arraystretch{1.2}
	\footnotesize
	\begin{center}
		\begin{tabular}
			{c|cc|cc}
			\toprule[1.5pt]
			Methods & Classifier & RMSE $\downarrow$ & Classifier & RMSE $\downarrow$ \\
			\midrule[1.5pt]
			ConfScore \cite{hendrycks2016baseline}& ResNet-56 & 6.985 & RepVGG-A0 & 8.722 \\ \hline
   Entropy \cite{guillory2021predicting} & ResNet-56 & 7.402& RepVGG-A0 & 9.093 \\ \hline
   Rotation \cite{deng2021does} & ResNet-56& 7.129& RepVGG-A0 & 13.391 \\ \hline
   ATC \cite{gargleveraging}& ResNet-56 & 7.766& RepVGG-A0 & 8.132 \\ \hline
   FID \cite{deng2021labels} & ResNet-56& 4.985& RepVGG-A0 & 5.966 \\ \hline
   AVG  & ResNet-56& 3.596& RepVGG-A0 & 4.244 \\ \hline
   OMFD (ours) & ResNet-56& \cellcolor[HTML]{FFCC67} 2.873& RepVGG-A0 & \cellcolor[HTML]{FFCC67}3.870 \\
			\bottomrule[1.3pt]
		\end{tabular}
	\end{center}
	\caption{The validation set results on mltiple autoeval methods. ``Entries'' indicates the number of submissions; $\downarrow$ means the smaller the value, the better.}
	\label{table:omfd}
\end{table}

The Table \ref{table:methods} provides us with some interesting conclusions. First, no single method can permanently lead on different validation sets, such as CIFAR-10.1, CIFAR-10.1-C, and CIFAR-10-F. This suggests that the unlabeled data have different domain shifts for diverse feature distributions. Second, overall, our proposed method KCFCA yields relatively robust results, including optimal performance on ``Dataset B'' and second-best performance on ``Dataset A''. 

\subsubsection{Experiments on various regression methods}
 \label{Experiments_regre}
Our intuition tells us that choosing a variety of regressors will lead to different performance impacts. To investigate this, we perform an exhaustive experimental comparison of two pre-trained ResNet-56 and RepVGG-A0 models, using LinearRegression (LR), KNeighborsRegressor (KNN), SVR, MLPRegressor (MLP), RandomForestRegressor (RFR), our proposed Dynamic Regression Model (DRM) as regression models on ConfScore \cite{hendrycksbaseline}, Entropy \cite{guillory2021predicting}, Rotation \cite{deng2021does}, ATC \cite{gargleveraging}, and FID \cite{deng2021labels}. LR, KNN, SVR, MLP, and RFR are provided by Scikit-learn.

As shown in Table \ref{table:regression}, it is rare to find a regression model that guarantees to outperform under all methods. However, it is evident that MLP has the least satisfactory outcome. It is encouraging to find that our proposed DRM can achieve relatively stable and excellent performance across different methods. For the final challenge submission, we experimentally use the LR regressor on ResNet-56 and the DRM regressor on RepVGG-A0.

\subsubsection{Experiments on mltiple autoeval methods}
\label{Experiments_methods}
To investigate the impact of different model fusion methods on the final challenge results, we conducted a series of experiments. The methods include ConfScore \cite{hendrycksbaseline}, Entropy \cite{guillory2021predicting}, Rotation \cite{deng2021does}, ATC \cite{gargleveraging}, FID \cite{deng2021labels}, the average of all methods (AVG), and our proposed Outlier Model Factor Discovery (OMFD). We present all the results in Table \ref{table:omfd}.

The table indicates that averaging the results of all methods leads to decent results, surpassing any single method, with an average score of 3.870. However, leveraging our OMFD method to eliminate anomalous outlier methods leads to surprising optimal results of 2.873, a 34.7\% improvement over the average score. Thus, our findings suggest that the inclusion of anomalous outlier methods is detrimental to the fusion process and adversely affects the final model output.

\begin{table}
\renewcommand\arraystretch{1.2}
	\footnotesize
	\begin{center}
		\begin{tabular}
			{c|c|c}
			\toprule[1.5pt]
			Teams & Classifier models  & RMSE $\downarrow$ \\
			\midrule[1.5pt]
			dlyldxwl & ResNet-56 \& RepVGG-A0 & 6.3746 \\ \hline
            \textbf{Yanglegeyang (ours)} & ResNet-56 \& RepVGG-A0 & \textbf{6.8526}\\ \hline
			SunshineBBB & ResNet-56 \& RepVGG-A0 & 6.9438 \\ \hline
                Shiny & ResNet-56 \& RepVGG-A0 & 8.6626 \\ \hline
                b136522541 & ResNet-56 \& RepVGG-A0 & 9.6994 \\ \hline
                xingjian & ResNet-56 \& RepVGG-A0 & 10.7378 \\ 
			\bottomrule[1.3pt]
		\end{tabular}
	\end{center}
	\caption{The test set results on the DataCV Challenge leaderboard. ``Entries'' indicates the number of submissions; $\downarrow$ means the smaller the value, the better.}
	\label{table:leaderboard}
\end{table}

\subsection{Results on DataCV Challenge}
In the challenge, there are two models ResNet-56 and RepVGG-
A0 is to be evaluated on the unlabeled test set in total by RMSE. The results of the challenge are shown in Table \ref{table:leaderboard}. 

For our final challenge submission, we combined the K-means Clustering Based Feature Consistency Alignment (KCFCA), Dynamic Regression Model (DRM), and Outlier Model Factor Discovery (OMFD) methods. Our team secured second place in the challenge, as shown in the table. Additionally, our proposed approach outperformed the optimal model results \cite{deng2021labels} provided in the challenge, achieving a 36\% improvement with a RMSE score of 6.8526 compared to 10.7378.

\section{Conclusion}
This paper highlights the various strategies we adopted in the challenge. Specifically, we propose the K-means Clustering Based Feature Consistency Alignment method to represent distribution shifts in different datasets, Dynamic Regression Model to analyze the relationship between shifts and model performance, and Outlier Model Factor Discovery to remove anomalous outlier autoeval models. Our approach secured second place in the challenge ranking. Furthermore, our KCFCA method achieved the most robust and optimal single model performance on the validation dataset.


{\small
\bibliographystyle{ieee_fullname}
\bibliography{egbib}

\begin{thebibliography}{10}\itemsep=-1pt

\bibitem{DataCVChallenge}
The 1st datacv challenge @ cvpr 2023.
\newblock In {\em https://sites.google.com/view/vdu-cvpr23/competition}.

\bibitem{arora2018stronger}
Sanjeev Arora, Rong Ge, Behnam Neyshabur, and Yi Zhang.
\newblock Stronger generalization bounds for deep nets via a compression
  approach.
\newblock In {\em International Conference on Machine Learning}, pages
  254--263. PMLR, 2018.

\bibitem{chen2018closing}
Jinghui Chen, Dongruo Zhou, Yiqi Tang, Ziyan Yang, Yuan Cao, and Quanquan Gu.
\newblock Closing the generalization gap of adaptive gradient methods in
  training deep neural networks.
\newblock {\em arXiv preprint arXiv:1806.06763}, 2018.

\bibitem{chen2020more}
Lin Chen, Yifei Min, Mingrui Zhang, and Amin Karbasi.
\newblock More data can expand the generalization gap between adversarially
  robust and standard models.
\newblock In {\em International Conference on Machine Learning}, pages
  1670--1680. PMLR, 2020.

\bibitem{chen2021mandoline}
Mayee Chen, Karan Goel, Nimit~S Sohoni, Fait Poms, Kayvon Fatahalian, and
  Christopher R{\'e}.
\newblock Mandoline: Model evaluation under distribution shift.
\newblock In {\em International Conference on Machine Learning}, pages
  1617--1629. PMLR, 2021.

\bibitem{corneanu2020computing}
Ciprian~A Corneanu, Sergio Escalera, and Aleix~M Martinez.
\newblock Computing the testing error without a testing set.
\newblock In {\em Proceedings of the IEEE/CVF Conference on Computer Vision and
  Pattern Recognition}, pages 2677--2685, 2020.

\bibitem{deng2009imagenet}
Jia Deng, Wei Dong, Richard Socher, Li-Jia Li, Kai Li, and Li Fei-Fei.
\newblock Imagenet: A large-scale hierarchical image database.
\newblock In {\em 2009 IEEE conference on computer vision and pattern
  recognition}, pages 248--255. Ieee, 2009.

\bibitem{deng2021does}
Weijian Deng, Stephen Gould, and Liang Zheng.
\newblock What does rotation prediction tell us about classifier accuracy under
  varying testing environments?
\newblock In {\em International Conference on Machine Learning}, pages
  2579--2589. PMLR, 2021.

\bibitem{deng2021labels}
Weijian Deng and Liang Zheng.
\newblock Are labels always necessary for classifier accuracy evaluation?
\newblock In {\em Proceedings of the IEEE/CVF Conference on Computer Vision and
  Pattern Recognition}, pages 15069--15078, 2021.

\bibitem{devries2018learning}
Terrance DeVries and Graham~W Taylor.
\newblock Learning confidence for out-of-distribution detection in neural
  networks.
\newblock {\em arXiv preprint arXiv:1802.04865}, 2018.

\bibitem{ding2021repvgg}
Xiaohan Ding, Xiangyu Zhang, Ningning Ma, Jungong Han, Guiguang Ding, and Jian
  Sun.
\newblock Repvgg: Making vgg-style convnets great again.
\newblock In {\em Proceedings of the IEEE/CVF conference on computer vision and
  pattern recognition}, pages 13733--13742, 2021.

\bibitem{dowson1982frechet}
DC Dowson and BV666017 Landau.
\newblock The fr{\'e}chet distance between multivariate normal distributions.
\newblock {\em Journal of multivariate analysis}, 12(3):450--455, 1982.

\bibitem{pascal-voc-2012}
M. Everingham, L. Van~Gool, C.~K.~I. Williams, J. Winn, and A. Zisserman.
\newblock The {PASCAL} {V}isual {O}bject {C}lasses {C}hallenge 2012 {(VOC2012)}
  {R}esults.
\newblock
  http://www.pascal-network.org/challenges/VOC/voc2012/workshop/index.html.

\bibitem{gargleveraging}
Saurabh Garg, Sivaraman Balakrishnan, Zachary~Chase Lipton, Behnam Neyshabur,
  and Hanie Sedghi.
\newblock Leveraging unlabeled data to predict out-of-distribution performance.
\newblock In {\em International Conference on Learning Representations}.

\bibitem{geifman2017selective}
Yonatan Geifman and Ran El-Yaniv.
\newblock Selective classification for deep neural networks.
\newblock {\em Advances in neural information processing systems}, 30, 2017.

\bibitem{guillory2021predicting}
Devin Guillory, Vaishaal Shankar, Sayna Ebrahimi, Trevor Darrell, and Ludwig
  Schmidt.
\newblock Predicting with confidence on unseen distributions.
\newblock In {\em Proceedings of the IEEE/CVF International Conference on
  Computer Vision}, pages 1134--1144, 2021.

\bibitem{hartigan1979algorithm}
John~A Hartigan and Manchek~A Wong.
\newblock Algorithm as 136: A k-means clustering algorithm.
\newblock {\em Journal of the royal statistical society. series c (applied
  statistics)}, 28(1):100--108, 1979.

\bibitem{hartigan1979k}
John~A Hartigan, Manchek~A Wong, et~al.
\newblock A k-means clustering algorithm.
\newblock {\em Applied statistics}, 28(1):100--108, 1979.

\bibitem{he2016deep}
Kaiming He, Xiangyu Zhang, Shaoqing Ren, and Jian Sun.
\newblock Deep residual learning for image recognition.
\newblock In {\em Proceedings of the IEEE conference on computer vision and
  pattern recognition}, pages 770--778, 2016.

\bibitem{hendrycks2019robustness}
Dan Hendrycks and Thomas Dietterich.
\newblock Benchmarking neural network robustness to common corruptions and
  perturbations.
\newblock {\em Proceedings of the International Conference on Learning
  Representations}, 2019.

\bibitem{hendrycksbaseline}
Dan Hendrycks and Kevin Gimpel.
\newblock A baseline for detecting misclassified and out-of-distribution
  examples in neural networks.
\newblock In {\em International Conference on Learning Representations}.

\bibitem{hendrycks2016baseline}
Dan Hendrycks and Kevin Gimpel.
\newblock A baseline for detecting misclassified and out-of-distribution
  examples in neural networks.
\newblock {\em arXiv preprint arXiv:1610.02136}, 2016.

\bibitem{DBLP:journals/corr/HuangLW16a}
Gao Huang, Zhuang Liu, and Kilian~Q. Weinberger.
\newblock Densely connected convolutional networks.
\newblock {\em CoRR}, abs/1608.06993, 2016.

\bibitem{ji2021predicting}
Xu Ji, Razvan Pascanu, R~Devon Hjelm, Andrea Vedaldi, Balaji Lakshminarayanan,
  and Yoshua Bengio.
\newblock Predicting unreliable predictions by shattering a neural network.
\newblock 2021.

\bibitem{jiang2018predicting}
Yiding Jiang, Dilip Krishnan, Hossein Mobahi, and Samy Bengio.
\newblock Predicting the generalization gap in deep networks with margin
  distributions.
\newblock {\em arXiv preprint arXiv:1810.00113}, 2018.

\bibitem{jiang2021assessing}
Yiding Jiang, Vaishnavh Nagarajan, Christina Baek, and J~Zico Kolter.
\newblock Assessing generalization of sgd via disagreement.
\newblock {\em arXiv preprint arXiv:2106.13799}, 2021.

\bibitem{kodinariya2013review}
Trupti~M Kodinariya, Prashant~R Makwana, et~al.
\newblock Review on determining number of cluster in k-means clustering.
\newblock {\em International Journal}, 1(6):90--95, 2013.

\bibitem{liang2017enhancing}
Shiyu Liang, Yixuan Li, and Rayadurgam Srikant.
\newblock Enhancing the reliability of out-of-distribution image detection in
  neural networks.
\newblock {\em arXiv preprint arXiv:1706.02690}, 2017.

\bibitem{likas2003global}
Aristidis Likas, Nikos Vlassis, and Jakob~J Verbeek.
\newblock The global k-means clustering algorithm.
\newblock {\em Pattern recognition}, 36(2):451--461, 2003.

\bibitem{lin2014microsoft}
Tsung-Yi Lin, Michael Maire, Serge Belongie, James Hays, Pietro Perona, Deva
  Ramanan, Piotr Doll{\'a}r, and C~Lawrence Zitnick.
\newblock Microsoft coco: Common objects in context.
\newblock In {\em Computer Vision--ECCV 2014: 13th European Conference, Zurich,
  Switzerland, September 6-12, 2014, Proceedings, Part V 13}, pages 740--755.
  Springer, 2014.

\bibitem{liu2020energy}
Weitang Liu, Xiaoyun Wang, John Owens, and Yixuan Li.
\newblock Energy-based out-of-distribution detection.
\newblock {\em Advances in neural information processing systems},
  33:21464--21475, 2020.

\bibitem{miao2022balanced}
Shuyu Miao, Shanshan Du, Rui Feng, Yuejie Zhang, Huayu Li, Tianbi Liu, Lin
  Zheng, and Weiguo Fan.
\newblock Balanced single-shot object detection using cross-context
  attention-guided network.
\newblock {\em Pattern Recognition}, 122:108258, 2022.

\bibitem{miao2021disentangled}
Shuyu Miao, Shuaicheng Li, Lin Zheng, Wei Yu, Jingjing Liu, Mingming Gong, and
  Rui Feng.
\newblock Disentangled feature network for fine-grained recognition.
\newblock In {\em Neural Information Processing: 28th International Conference,
  ICONIP 2021, Sanur, Bali, Indonesia, December 8--12, 2021, Proceedings, Part
  II 28}, pages 439--450. Springer, 2021.

\bibitem{na2010research}
Shi Na, Liu Xumin, and Guan Yong.
\newblock Research on k-means clustering algorithm: An improved k-means
  clustering algorithm.
\newblock In {\em 2010 Third International Symposium on intelligent information
  technology and security informatics}, pages 63--67. Ieee, 2010.

\bibitem{neyshabur2017exploring}
Behnam Neyshabur, Srinadh Bhojanapalli, David McAllester, and Nati Srebro.
\newblock Exploring generalization in deep learning.
\newblock {\em Advances in neural information processing systems}, 30, 2017.

\bibitem{ovadia2019can}
Yaniv Ovadia, Emily Fertig, Jie Ren, Zachary Nado, David Sculley, Sebastian
  Nowozin, Joshua Dillon, Balaji Lakshminarayanan, and Jasper Snoek.
\newblock Can you trust your model's uncertainty? evaluating predictive
  uncertainty under dataset shift.
\newblock {\em Advances in neural information processing systems}, 32, 2019.

\bibitem{pham2005selection}
Duc~Truong Pham, Stefan~S Dimov, and Chi~D Nguyen.
\newblock Selection of k in k-means clustering.
\newblock {\em Proceedings of the Institution of Mechanical Engineers, Part C:
  Journal of Mechanical Engineering Science}, 219(1):103--119, 2005.

\bibitem{recht2018cifar10.1}
Benjamin Recht, Rebecca Roelofs, Ludwig Schmidt, and Vaishaal Shankar.
\newblock Do cifar-10 classifiers generalize to cifar-10?
\newblock 2018.
\newblock \url{https://arxiv.org/abs/1806.00451}.

\bibitem{ren2019likelihood}
Jie Ren, Peter~J Liu, Emily Fertig, Jasper Snoek, Ryan Poplin, Mark Depristo,
  Joshua Dillon, and Balaji Lakshminarayanan.
\newblock Likelihood ratios for out-of-distribution detection.
\newblock {\em Advances in neural information processing systems}, 32, 2019.

\bibitem{schiff2021predicting}
Yair Schiff, Brian Quanz, Payel Das, and Pin-Yu Chen.
\newblock Predicting deep neural network generalization with perturbation
  response curves.
\newblock {\em Advances in Neural Information Processing Systems},
  34:21176--21188, 2021.

\bibitem{NEURIPS2021_01894d6f}
Yiyou Sun, Chuan Guo, and Yixuan Li.
\newblock React: Out-of-distribution detection with rectified activations.
\newblock In M. Ranzato, A. Beygelzimer, Y. Dauphin, P.S. Liang, and J.~Wortman
  Vaughan, editors, {\em Advances in Neural Information Processing Systems},
  volume~34, pages 144--157. Curran Associates, Inc., 2021.

\bibitem{torralba2008tinyimages}
Antonio Torralba, Rob Fergus, and William~T. Freeman.
\newblock 80 million tiny images: A large data set for nonparametric object and
  scene recognition.
\newblock {\em IEEE Transactions on Pattern Analysis and Machine Intelligence},
  30(11):1958--1970, 2008.

\bibitem{pmlr-v119-yang20j}
Zitong Yang, Yaodong Yu, Chong You, Jacob Steinhardt, and Yi Ma.
\newblock Rethinking bias-variance trade-off for generalization of neural
  networks.
\newblock In Hal~Daumé III and Aarti Singh, editors, {\em Proceedings of the
  37th International Conference on Machine Learning}, volume 119 of {\em
  Proceedings of Machine Learning Research}, pages 10767--10777. PMLR, 13--18
  Jul 2020.

\bibitem{yu2022predicting}
Yaodong Yu, Zitong Yang, Alexander Wei, Yi Ma, and Jacob Steinhardt.
\newblock Predicting out-of-distribution error with the projection norm.
\newblock In {\em International Conference on Machine Learning}, pages
  25721--25746. PMLR, 2022.

\end{thebibliography}
}

\end{document}